\documentclass{article}

\PassOptionsToPackage{numbers, compress}{natbib}

\usepackage[preprint]{neurips_2026}

\usepackage[utf8]{inputenc}
\usepackage[T1]{fontenc}
\usepackage[hidelinks]{hyperref}
\usepackage{url}
\usepackage{booktabs}
\usepackage{amsfonts}
\usepackage{nicefrac}
\usepackage{microtype}
\usepackage{xcolor}
\usepackage[pdftex]{graphicx}
\usepackage{subcaption}
\usepackage{wrapfig}
\graphicspath{{./img/}}
\usepackage{multirow}
\usepackage{enumitem}
\usepackage{xcolor}
\usepackage{amsmath}

\title{PRISM: Iterative Cross-Modal Posterior Refinement for Dynamic Text-Attributed Graphs}

\author{
Trimble Chang \quad Yihang Liu \quad Mingjing Han\thanks{Corresponding Author} \quad Han Zhang \\
College of Artificial Intelligence \\
Nankai University \\
\texttt{\{trimblechang, yihangliu, hanmj\}@mail.nankai.edu.cn} \\
\texttt{zhanghan@nankai.edu.cn}
}

\begin{document}

\maketitle

\begin{abstract}
  Dynamic text-attributed graphs (DyTAGs) provide a powerful framework for modeling evolving systems in which node semantics and time-dependent interactions are tightly coupled. Recently, multimodal learning has emerged as a promising yet underexplored direction for enhancing DyTAG representation learning. However, existing methods typically rely on rigid modality partitions and one-shot fusion strategies, which limit their ability to capture the intrinsic and evolving dependencies between node semantics and interaction behaviors. To address these limitations, we propose \textbf{PRISM}, an iterative cross-modal posterior refinement framework for DyTAG representation learning. PRISM organizes DyTAG information into semantic and behavioral modalities, providing a more intrinsic alternative to carrier-level modality partitions. Instead of fusing the two modalities in a single step, PRISM learns a refinement trajectory that progressively transforms semantic priors into behavior-conditioned posterior states through cross-modal interaction with behavioral evidence. Extensive experiments on DTGB benchmark datasets show that PRISM achieves strong performance on temporal link prediction and destination node retrieval tasks. Further ablation studies validate the effectiveness of semantic--behavioral modeling and iterative posterior refinement.
\end{abstract}

\section{Introduction}
\label{sec:introduction}

Dynamic graphs provide a fundamental abstraction for modeling evolving systems~\cite{kazemi2020representation,skarding2021foundations,feng2025comprehensive}, such as social networks~\cite{leskovec2008microscopic,huang2022ttergm,sun2022aligning}, recommendation systems~\cite{he2020lightgcn,zhang2022dynamic,tang2023dynamic}, and temporal knowledge graphs~\cite{jin2020recurrent,li2021temporal,cai2023temporal}. In many real-world scenarios, interactions are also accompanied by textual information, giving rise to dynamic text-attributed graphs (DyTAGs)~\cite{zhang2024dtgb,peng2026gdgb}. Unlike commonly studied TAGs~\cite{hu2020open,he2023explanations,yan2023comprehensive} and CTDGs~\cite{nguyen2018continuous,huang2023temporal,gastinger2024tgb}, DyTAGs require models to capture the coupled evolution between semantic attributes and dynamic behaviors. For example, in e-commerce review graphs, textual profiles or descriptions provide semantic anchors, while time-stamped reviews continuously reshape user preferences and product perception. Therefore, effective DyTAG learning requires representations that jointly reason over node semantics and historical behavioral context.

Existing DyTAG modeling approaches mainly follow two lines: TGNN-based temporal graph modeling and LLM-based semantic reasoning. TGNN methods usually encode node or edge texts into vector features and then apply temporal graph encoders to model evolving interactions~\cite{kumar2019jodie,trivedi2019dyrep,xu2020tgat,rossi2020tgn,wang2021cawn,wang2021tcl,cong2023graphmixer,yu2023dygformer}. While effective for temporal structures and local interaction patterns, they often treat textual information as initialization features or auxiliary attributes, leaving node-internal semantics and higher-level behavioral regularities underexplored. LLM-based methods leverage large language models to process temporal neighborhoods, interaction texts, or evolving node descriptions~\cite{roy2025lkd4dytag,zhang2025cross,wang2026dygrasp}. These methods strengthen semantic modeling but introduce substantial computational overhead and often rely on explicit textual reasoning pipelines, making it difficult to obtain lightweight and unified time-dependent node states.

More recently, multimodal learning has emerged as a promising direction for representation learning, with prior studies emphasizing modality interaction, alignment, and fusion~\cite{baltrusaitis2019multimodal,zadeh2017tensor,tsai2019multimodal}. In multimodal knowledge graphs, incorporating non-structural signals such as images or textual descriptions can enrich entity representations and improve knowledge graph completion~\cite{xie2017ikrl,chen2022mkgformer,liu2024multimodalkgc}. For DyTAGs, recent work has also made an initial attempt to explicitly model and integrate textual, temporal, and structural modalities for dynamic graph representation learning~\cite{xu2026moment}. However, existing multimodal DyTAG methods typically rely on carrier-level modality partitions followed by one-shot fusion. This design overlooks a more intrinsic distinction: node text provides relatively stable semantic identity, whereas historical interactions provide dynamic behavioral evidence. A single fusion step may therefore be insufficient for modeling their time-varying dependency.

\begin{figure}[htbp]
  \centering
  \includegraphics[width=1.0\columnwidth]{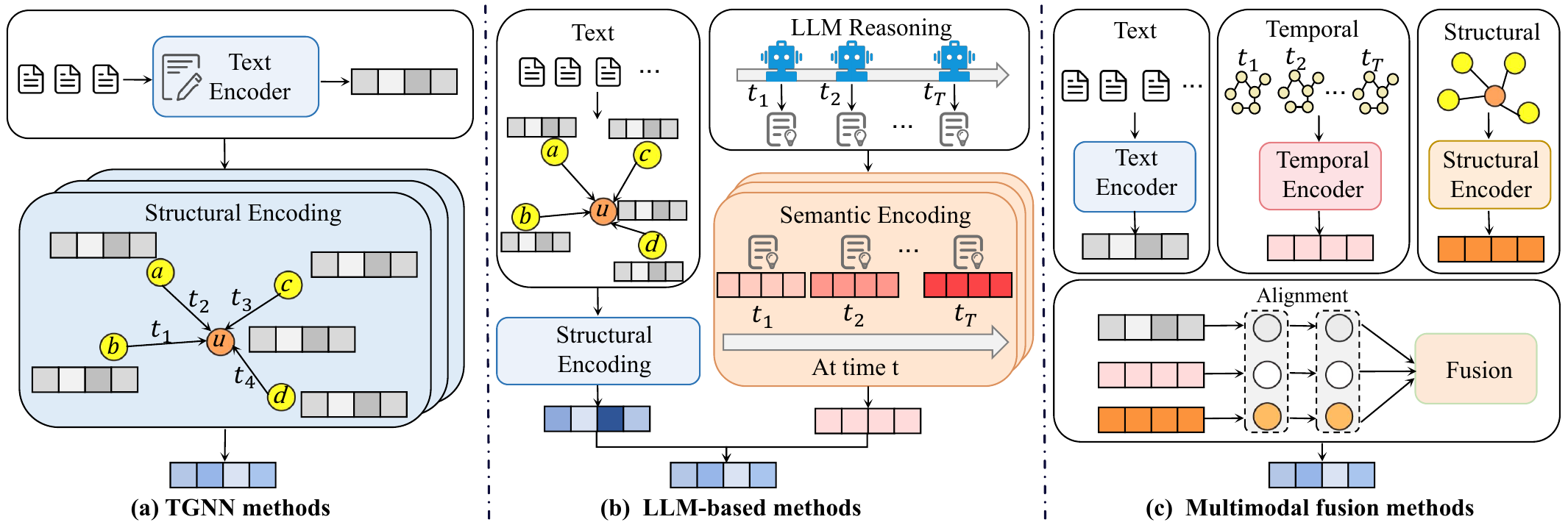}
  \caption{Comparison of existing DyTAG modeling paradigms. (a) TGNN-based methods encode textual attributes as input features and model temporal interactions with temporal graph encoders. (b) LLM-based methods leverage language models to reason over textual attributes and interaction contexts. (c) Multimodal learning methods encode heterogeneous information sources with modality-specific modules and integrate them for representation learning.}
  \label{fig:intro}
\end{figure}

To address these limitations, we propose \textbf{PRISM} (\textbf{P}osterior \textbf{R}efinement via \textbf{I}terative \textbf{S}emantic-behavioral \textbf{M}odulation), an iterative cross-modal posterior refinement framework for DyTAG representation learning. PRISM organizes DyTAG information into semantic and behavioral modalities, describing node states from complementary views. The semantic modality acts as a prior description of node identity, while the behavioral modality serves as dynamic evidence of the node's evolving context. From this perspective, multimodal fusion becomes a posterior inference process rather than a one-shot feature combination. Specifically, PRISM starts from semantic prior states and repeatedly retrieves behavioral evidence through state-conditioned cross-modal attention. Inspired by iterative state refinement in diffusion and flow-based modeling, PRISM learns task-driven refinement directions and applies Euler-style updates to progressively transform semantic priors into behavior-conditioned posterior states. This design preserves semantic identity while allowing node representations to adapt to time-dependent behavioral contexts.

The main contributions of our work are summarized as follows:
\begin{itemize}[leftmargin=15pt, itemsep=0pt]
    \item We propose a semantic--behavioral perspective for DyTAG multimodal learning, which distinguishes stable semantic identity from time-dependent behavioral evidence and provides a more intrinsic alternative to carrier-level modality partition.
    \item We develop PRISM, an iterative cross-modal posterior refinement framework that treats multimodal fusion as a prior-to-posterior state refinement process. PRISM learns state-conditioned refinement directions to progressively transform semantic priors into behavior-conditioned posterior node states.
    \item We conduct extensive experiments on benchmark DyTAG datasets, showing that PRISM achieves strong performance, with further analyses validating the effectiveness of semantic--behavioral modeling and iterative refinement.
\end{itemize}

\section{Preliminaries}
\label{sec:preliminaries}

\paragraph{Dynamic text-attributed graphs.}
A DyTAG is a continuous-time interaction graph in which both nodes and temporal interactions are associated with textual attributes. We denote a DyTAG as a chronological sequence of interactions
\begin{equation}
    \mathcal{G}
    =
    \{(u_i,v_i,r_i,t_i)\}_{i=1}^{|\mathcal{E}|},
    \quad
    t_1 \leq t_2 \leq \cdots \leq t_{|\mathcal{E}|},
    \label{eq:dytag_definition}
\end{equation}
where $u_i,v_i\in\mathcal{V}$ denote the source and destination nodes, $r_i\in\mathcal{R}$ denotes the textual attribute associated with the interaction, and $t_i\in\mathbb{R}_{+}$ is the timestamp. Each node $u\in\mathcal{V}$ is also associated with a textual attribute $d_u\in\mathcal{D}$. We use $\mathcal{G}_{<t}$ to denote the subgraph consisting of all interactions that occur before timestamp $t$.

\paragraph{Representation learning on DyTAGs.}
Given a DyTAG $\mathcal{G}$ and a candidate interaction $(u,v,t)$, representation learning on DyTAGs aims to learn a time-dependent encoder
\begin{equation}
    F_{\theta}:
    (u,v,t,\mathcal{G}_{<t})
    \mapsto
    \mathbf{z}_{u}^{t},\mathbf{z}_{v}^{t}
    \in\mathbb{R}^{d},
    \label{eq:representation_function}
\end{equation}
where $\mathbf{z}_{u}^{t}$ and $\mathbf{z}_{v}^{t}$ are the representations of nodes $u$ and $v$ at time $t$ conditioned on the historical graph $\mathcal{G}_{<t}$. The learned representations are then fed into a task-specific decoder to estimate the likelihood of the candidate interaction. For link prediction, the model is optimized to assign higher likelihoods to observed future interactions than to negative candidates; for destination node retrieval, candidate nodes are ranked according to their decoded interaction likelihoods with the source node at time $t$.

\section{Related Work}
\label{sec:related}

\paragraph{TGNN methods for DyTAG learning.} Continuous-time dynamic graph learning has been widely studied for modeling time-stamped interactions. Early memory-based methods, such as JODIE, DyRep, and TGN, maintain evolving node states through recurrent memory updates as new events arrive~\cite{kumar2019jodie,trivedi2019dyrep,rossi2020tgn}. Memory-free methods instead compute temporal representations from historical neighborhoods, including attention-based models such as TGAT, CAWN, and TCL~\cite{xu2020tgat,wang2021cawn,wang2021tcl}, as well as lightweight sequence-based architectures such as GraphMixer and DyGFormer~\cite{cong2023graphmixer,yu2023dygformer}. These methods provide strong temporal graph backbones and are commonly adopted in DyTAG benchmarks~\cite{zhang2024dtgb}. However, when applied to DyTAGs, they typically encode textual attributes as static node or edge features and then inject them into structure-centered temporal encoders. Such a design is effective for modeling local interaction patterns, but it often treats textual semantics as auxiliary signals, leaving node-internal semantics and higher-level behavioral regularities insufficiently explored.

\paragraph{LLM-based methods for DyTAG learning.} Large language models have recently been introduced to DyTAG learning due to their strong ability to process textual attributes and interaction contexts. LKD4DyTAG distills knowledge from LLM-driven edge representations into a lightweight spatio-temporal GNN~\cite{roy2025lkd4dytag}, while CROSS uses LLMs to extract dynamic textual semantics and integrates them with evolving graph structures through a semantic-structural co-encoder~\cite{zhang2025cross}. DyGRASP further captures recent and global temporal semantics through node-centric implicit reasoning and explicit reasoning chains, and then combines the resulting semantic features with temporal GNN representations~\cite{wang2026dygrasp}. These methods demonstrate the value of semantic reasoning for DyTAGs, but they often rely on prompt construction, long-context processing, repeated summarization, or teacher-student distillation, which may introduce considerable computational overhead. Moreover, semantic reasoning and temporal graph representation are usually connected through feature injection or merging, rather than a unified state inference process. In contrast, PRISM aims to learn lightweight posterior node states by iteratively refining semantic priors with behavioral evidence.

\paragraph{Multimodal fusion methods for DyTAG learning.} Multimodal learning seeks to exploit complementary signals from heterogeneous sources and has been broadly studied in representation learning~\cite{baltrusaitis2019multimodal,tsai2019multimodal}. In graph learning, multimodal knowledge graph methods have shown that incorporating non-structural information, such as images or textual descriptions, can enrich entity representations and improve downstream reasoning~\cite{xie2017ikrl}. For DyTAGs, MoMent makes an initial attempt to explicitly model textual, temporal, and structural modalities with modality-specific encoders and align their latent spaces for dynamic graph representation learning~\cite{xu2026moment}. Despite its promise, existing multimodal DyTAG learning still largely follows carrier-level modality partitions and typically performs fusion after separate modality encoding. This paradigm emphasizes modality aggregation, but does not fully capture the intrinsic dependency between semantic identity and dynamic behavioral context. PRISM departs from this view by organizing DyTAG information into semantic and behavioral modalities, where node text provides a semantic prior and historical interactions provide behavioral evidence. Multimodal fusion is therefore formulated as iterative posterior refinement rather than one-shot feature aggregation.

\section{Method}
\label{sec:method}

\begin{figure}[htbp]
  \centering
  \includegraphics[width=1.0\columnwidth]{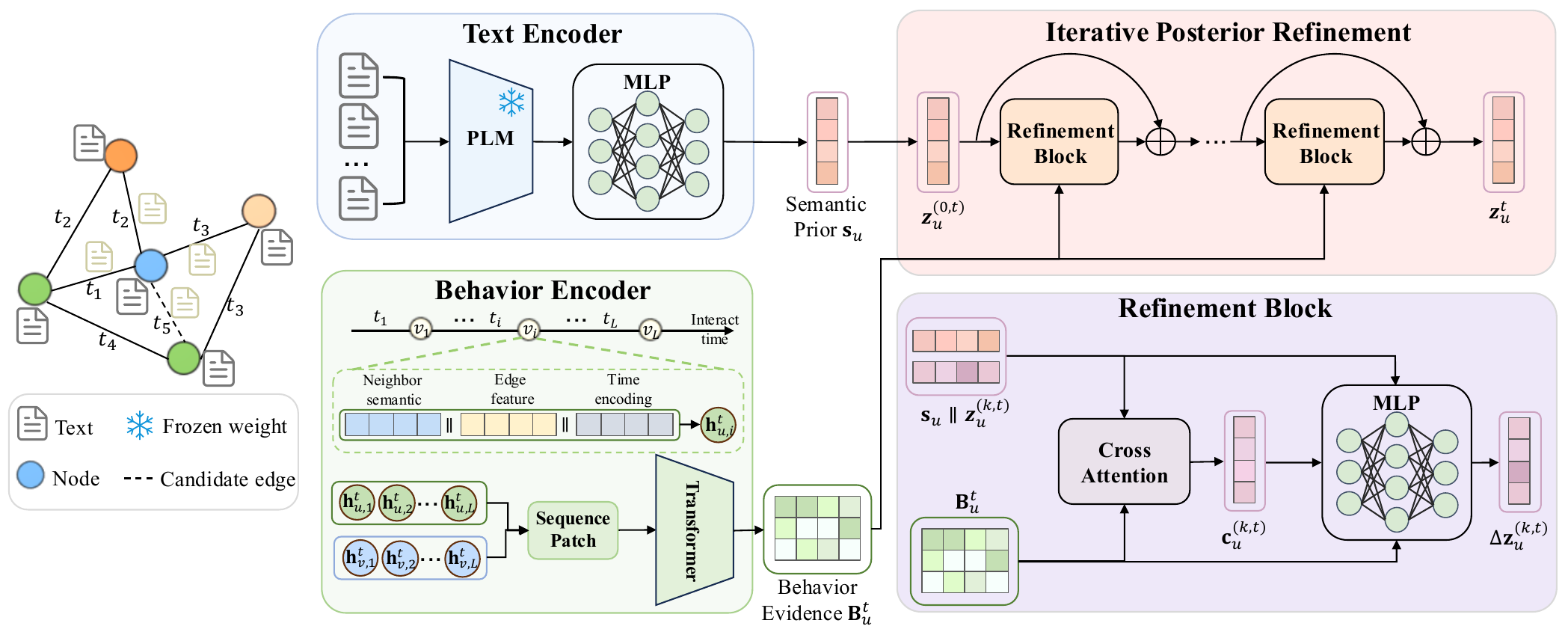}
  \caption{Overview of PRISM. PRISM encodes textual attributes as semantic priors and extracts historical interactions as behavioral evidence. It then performs iterative cross-modal posterior refinement to progressively transform semantic priors into behavior-conditioned posterior states.}
  \label{fig:model_arch}
\end{figure}

In this section, we present the overall framework of PRISM. As illustrated in Figure~\ref{fig:model_arch}, PRISM first organizes DyTAG information into semantic and behavioral modalities, then performs iterative cross-modal posterior refinement to transform semantic priors into time-dependent posterior states, and finally optimizes the refined states for downstream prediction tasks.

\subsection{Multimodal Modeling}
\label{subsec:multimodal_modeling}

\paragraph{Semantic Modal Modeling.}
For each node $u$, PRISM encodes its textual attribute $d_u$ into the semantic modality. Specifically, we first use a lightweight pre-trained language model to obtain the textual representation, and then project it into the model latent space:
\begin{equation}
    \mathbf{s}_u
    =
    \operatorname{FFN}
    \left(
        \operatorname{PLM}(d_u)
    \right),
    \label{eq:textual_node_encoding}
\end{equation}
where $\operatorname{PLM}(\cdot)$ is instantiated as BERT~\cite{devlin2019bert}, and $\operatorname{FFN}(\cdot)$ projects textual features into the latent space. The resulting $\mathbf{s}_u\in\mathbb{R}^{d}$ serves as the semantic prior of node $u$.

\paragraph{Behavioral Modal Modeling.}
The behavioral modality is constructed from historical interactions before the prediction time. For a candidate interaction $(u,v,t)$, PRISM collects the recent histories of both the source node $u$ and the candidate destination node $v$, so that behavioral modeling can capture node-specific temporal patterns as well as candidate-conditioned interaction compatibility. For node $u$, let
\begin{equation}
    \mathcal{H}_{u}^{t}
    =
    \{(v_i,r_{uv_i},t_i)\mid t_i<t\}_{i=1}^{L}
    \label{eq:history_sequence}
\end{equation}
denote its chronologically ordered historical interaction sequence, where $v_i$ is the historical neighbor, $r_{uv_i}$ is the edge text, and $t_i$ is the interaction timestamp. We keep the most recent $L$ interactions for each node and pad sequences with fewer than $L$ historical interactions; padded positions are excluded from the subsequent attention and pooling operations.

For each historical interaction, PRISM forms a behavioral token by combining the neighbor semantic representation, the edge semantic representation, and the relative time encoding:
\begin{equation}
    \mathbf{h}_{u,i}^{t}
    =
    \operatorname{FFN}
    \left(\left[
        \mathbf{s}_{v_i}
        \,\Vert\,
        \mathbf{e}_{uv_i}^{t_i}
        \,\Vert\,
        \phi(t-t_i)
    \right]\right),
    \label{eq:behavior_token}
\end{equation}
where $\mathbf{e}_{uv_i}^{t_i}$ denotes the encoded representation of the edge text, $\phi(\cdot)$ is a learnable time encoding function~\cite{xu2020tgat}, and $\Vert$ denotes concatenation. The token $\mathbf{h}_{u,i}^{t}$ represents a behavioral observation that records which neighbor was involved, what textual interaction occurred, and how far this event is from the current prediction time. We then organize these tokens into the behavioral sequence $\mathbf{H}_{u}^{t} = [\mathbf{h}_{u,1}^{t},\mathbf{h}_{u,2}^{t},\ldots,\mathbf{h}_{u,L}^{t}]$. The behavioral sequence $\mathbf{H}_{v}^{t}$ for the candidate destination node $v$ is constructed in the same way.

To obtain candidate-conditioned behavioral evidence, PRISM jointly encodes the two behavioral sequences with a Transformer encoder~\cite{vaswani2017attention} and then summarizes the encoded tokens into compact behavioral representations:
\begin{gather}
    [\mathbf{B}_{u}^{t},\mathbf{B}_{v}^{t}]
    =
    \operatorname{TransformerEncoder}
    \left(
        [\mathbf{H}_{u}^{t}\Vert\mathbf{H}_{v}^{t}]
    \right),
    \label{eq:behavior_transformer} \\
    \mathbf{b}_{u}^{t}
    =
    \operatorname{Pooling}
    \left(
        \mathbf{B}_{u}^{t}
    \right),
    \quad
    \mathbf{b}_{v}^{t}
    =
    \operatorname{Pooling}
    \left(
        \mathbf{B}_{v}^{t}
    \right).
    \label{eq:behavior_readout}
\end{gather}
The Transformer output is split back into source-side and destination-side behavioral observation tokens. Through this pair encoding, each node's historical behavior is interpreted in the context of the candidate counterpart, allowing the behavioral modality to capture recent interaction contexts, temporal regularities, and candidate-conditioned compatibility.

\subsection{Iterative Cross-Modal Posterior Refinement}
\label{subsec:posterior_refinement}

The semantic and behavioral modalities provide complementary but asymmetric information: text describes node identity, while historical interactions reveal how this identity is expressed under time-dependent contexts. Therefore, the goal of multimodal fusion is not merely to combine two feature vectors, but to infer a task-relevant node state conditioned on both semantic identity and historical evidence. PRISM formulates this process as an iterative posterior refinement problem: starting from a semantic prior, the model progressively incorporates behavioral evidence and approaches a time-dependent posterior state.

This formulation is inspired by inference-based multimodal fusion~\cite{wu2018multimodal,shi2019variational,sutter2020multimodal}, but PRISM does not rely on an explicit probabilistic inference framework. In DyTAGs, the true posterior node state is not directly observed, and specifying a tractable likelihood between textual semantics, temporal behaviors, and downstream links is difficult. Instead, we adopt a posterior-inspired latent state view. For a node $u$ at time $t$, PRISM aims to learn
\begin{equation}
    \mathbf{z}_u^t
    \approx
    F_\theta(\mathbf{s}_u, \mathbf{B}_u^t),
    \label{eq:posterior_mapping}
\end{equation}
where $\mathbf{s}_u$ acts as the semantic prior and $\mathbf{B}_u^t$ denotes behavioral evidence extracted from historical interactions before time $t$.

Rather than using a single fusion operator, which may prematurely compress heterogeneous evidence, PRISM parameterizes $F_\theta$ as an iterative refinement process. This design is inspired by diffusion and flow-based models~\cite{chen2018neural,ho2020denoising,lipman2023flow,liu2023flow}, where states are transformed through a sequence of small updates. However, PRISM does not generate node states from noise: the semantic representation $\mathbf{s}_u$ already contains informative identity-level prior knowledge. We therefore initialize refinement from the semantic prior:
\begin{equation}
    \mathbf{z}_{u}^{(0,t)} = \mathbf{s}_u .
    \label{eq:posterior_initial_state}
\end{equation}

We view the refinement process as a discretized trajectory from the semantic prior to the posterior state. At each step, PRISM estimates a state-dependent refinement direction conditioned on the current state, the semantic prior, and behavioral observations:
\begin{equation}
    \Delta \mathbf{z}_{u}^{(k,t)}
    =
    v_{\theta_k}
    \left(
        \mathbf{z}_{u}^{(k,t)},
        \mathbf{s}_u,
        \mathbf{B}_{u}^{t},
        \mathbf{b}_{u}^{t}
    \right),
    \label{eq:velocity_field_general}
\end{equation}
where $v_{\theta_k}(\cdot)$ can be interpreted as the learned cross-modal velocity field at the $k$-th refinement step. The final posterior representation is obtained by applying $K$ Euler-style refinement steps:
\begin{equation}
\begin{gathered}
    \mathbf{z}_{u}^{t}
    =
    \mathbf{z}_{u}^{(K,t)}, \\
    \mathbf{z}_{u}^{(k+1,t)}
    =
    \mathbf{z}_{u}^{(k,t)}
    +
    \frac{1}{K}
    \Delta \mathbf{z}_{u}^{(k,t)}.
\end{gathered}
\label{eq:euler_refinement}
\end{equation}

The factor $1/K$ normalizes the accumulated refinement magnitude, making the process comparable across different numbers of refinement steps. This Euler-style update turns multimodal fusion into a gradual state transition rather than an abrupt feature aggregation.

We instantiate the velocity field through state-conditioned cross-modal evidence retrieval. At the $k$-th step, the current posterior state forms a query together with the original semantic prior:
\begin{equation}
\begin{gathered}
    \mathbf{q}_{u}^{(k,t)}
    =
    W_q^{(k)}
    \left[
        \mathbf{z}_{u}^{(k,t)}
        \,\Vert\,
        \mathbf{s}_u
    \right], \\
    \mathbf{c}_{u}^{(k,t)}
    =
    \operatorname{Attn}
    \left(
        \mathbf{q}_{u}^{(k,t)},
        \mathbf{B}_{u}^{t},
        \mathbf{B}_{u}^{t}
    \right),
\end{gathered}
\label{eq:posterior_attention}
\end{equation}
where $\mathbf{c}_{u}^{(k,t)}$ denotes the behavioral evidence retrieved at the current refinement stage. Since the query depends on the evolving posterior state, different refinement steps can focus on different behavioral observations without requiring an explicit step embedding. Early steps may capture coarse behavioral tendencies, while later steps can perform more fine-grained corrections conditioned on the already refined state.

The retrieved behavioral context is then used to estimate the refinement velocity:
\begin{equation}
    \Delta\mathbf{z}_{u}^{(k,t)}
    =
    \operatorname{FFN}_{v}^{(k)}
    \left(
        \left[
            \mathbf{z}_{u}^{(k,t)}
            \,\Vert\,
            \mathbf{s}_u
            \,\Vert\,
            \mathbf{b}_{u}^{t}
            \,\Vert\,
            \mathbf{c}_{u}^{(k,t)}
        \right]
    \right).
    \label{eq:posterior_delta}
\end{equation}
Here, $\mathbf{b}_{u}^{t}$ provides a global summary of historical behavior, whereas $\mathbf{c}_{u}^{(k,t)}$ provides step-specific evidence retrieved from behavioral tokens. Their combination allows the refinement direction to be both globally behavior-aware and locally adaptive to the current state.

This design gives PRISM a conditional prior-to-posterior state transport view. Unlike one-shot fusion, PRISM learns how semantic states should move under behavioral evidence; unlike diffusion-based generation, it starts from informative text-derived priors rather than noise. The final representation thus preserves semantic identity while adapting to the behavioral context at time $t$.

\subsection{Training Objective}
\label{subsec:training_objective}

PRISM is trained end-to-end with a task-dependent supervised objective and three lightweight regularizers that preserve behavioral information, anchor semantic identity, and smooth the refinement trajectory.

For temporal link prediction, given a candidate interaction $(u,v,t)$, PRISM first obtains the posterior states $\mathbf{z}_{u}^{t}$ and $\mathbf{z}_{v}^{t}$ through the refinement process above. The interaction likelihood is then estimated by a prediction decoder:
\begin{equation}
    \hat{y}_{uv}^{t}
    =
    \sigma
    \left(
        \operatorname{MLP}
        \left(
            [\mathbf{z}_{u}^{t}
            \,\Vert\,
            \mathbf{z}_{v}^{t}
            ]
        \right)
    \right).
    \label{eq:prediction_score}
\end{equation}
Given a mini-batch $\mathcal{B}$ of observed temporal interactions, we sample a negative destination $v^{-}$ for each positive interaction $(u,v,t)$ and optimize the binary prediction loss:
\begin{equation}
    \mathcal{L}_{\mathrm{task}}
    =
    -
    \frac{1}{|\mathcal{B}|}
    \sum_{(u,v,t)\in\mathcal{B}}
    \left[
        \log \hat{y}_{uv}^{t}
        +
        \log
        \left(
            1-\hat{y}_{uv^{-}}^{t}
        \right)
    \right].
    \label{eq:task_loss}
\end{equation}
For other DyTAG tasks, such as node classification or interaction classification, $\mathcal{L}_{\mathrm{task}}$ can be replaced with the corresponding supervised loss while keeping the posterior refinement module unchanged.

Since posterior refinement is expected to incorporate behavioral evidence without destroying the semantic identity encoded by the prior, we introduce three auxiliary objectives. Let $\mathcal{V}_{h}$ denote the set of node-time instances with non-empty histories in the mini-batch. The first objective encourages the final posterior state to preserve behavior-relevant information by reconstructing the detached behavioral representation:
\begin{equation}
    \mathcal{L}_{\mathrm{recon}}
    =
    \frac{1}{|\mathcal{V}_{h}|}
    \sum_{(i,t)\in\mathcal{V}_{h}}
    \left\|
        \operatorname{MLP}_{\mathrm{recon}}(\mathbf{z}_{i}^{t})
        -
        \operatorname{sg}(\mathbf{b}_{i}^{t})
    \right\|_{2}^{2},
    \label{eq:behavior_reconstruction_loss}
\end{equation}
where $\operatorname{sg}(\cdot)$ denotes the stop-gradient operation. This objective does not force the posterior state to collapse into the behavioral representation. Instead, it provides a weak information-preserving constraint, requiring the posterior state to retain sufficient behavioral evidence after refinement.

The second objective prevents the posterior state from drifting excessively away from its semantic prior:
\begin{equation}
    \mathcal{L}_{\mathrm{margin}}
    =
    \frac{1}{|\mathcal{V}_{h}|}
    \sum_{(i,t)\in\mathcal{V}_{h}}
    \max
    \left(
        0,
        \left\|
            \mathbf{z}_{i}^{t}
            -
            \mathbf{s}_{i}
        \right\|_{2}^{2}
        -
        m
    \right),
    \label{eq:margin_loss}
\end{equation}
where $m$ is a predefined margin. This term acts as a soft trust-region constraint around the semantic prior. It allows the posterior state to move within a reasonable neighborhood of the semantic identity, but penalizes excessive deviations that may overfit noisy or sparse behavioral observations.

The third objective regularizes the refinement trajectory by discouraging excessively large updates between adjacent refinement steps:
\begin{equation}
    \mathcal{L}_{\mathrm{step}}
    =
    \frac{1}{K|\mathcal{V}_{h}|}
    \sum_{(i,t)\in\mathcal{V}_{h}}
    \sum_{k=0}^{K-1}
    \left\|
        \mathbf{z}_{i}^{(k+1,t)}
        -
        \mathbf{z}_{i}^{(k,t)}
    \right\|_{2}^{2}.
    \label{eq:step_regularization_loss}
\end{equation}
This term encourages the prior-to-posterior transition to proceed through stable residual corrections rather than abrupt representation shifts, which is consistent with the Euler-style refinement interpretation.

The final training objective is
\begin{equation}
    \mathcal{L}
    =
    \mathcal{L}_{\mathrm{task}}
    +
    \lambda_{\mathrm{recon}}\mathcal{L}_{\mathrm{recon}}
    +
    \lambda_{\mathrm{margin}}\mathcal{L}_{\mathrm{margin}}
    +
    \lambda_{\mathrm{step}}\mathcal{L}_{\mathrm{step}},
    \label{eq:overall_loss}
\end{equation}
where $\lambda_{\mathrm{recon}}$, $\lambda_{\mathrm{margin}}$, and $\lambda_{\mathrm{step}}$ control the strengths of the auxiliary objectives. The task loss ensures that the posterior states are optimized for the downstream prediction target, while the auxiliary objectives encourage the learned posterior states to be behavior-aware, semantically anchored, and smoothly refined across iterative updates.

\section{Experiments}
\label{sec:experiments}

\subsection{Experimental Settings}

We evaluate PRISM on eight datasets from the DTGB benchmark~\cite{zhang2024dtgb}: Enron, ICEWS1819, Googlemap CT, GDELT, Stack elec, Stack ubuntu, Amazon movies, and Yelp, which span diverse domains such as communication networks, event graphs, online communities, e-commerce, and local services. We compare PRISM with representative temporal graph learning and DyTAG modeling baselines, including JODIE~\cite{kumar2019jodie}, DyRep~\cite{trivedi2019dyrep}, TGAT~\cite{xu2020tgat}, CAWN~\cite{wang2021cawn}, TCL~\cite{wang2021tcl}, GraphMixer~\cite{cong2023graphmixer}, DyGFormer~\cite{yu2023dygformer}, and MoMent~\cite{xu2026moment}. Following the DTGB protocol, we evaluate models on temporal link prediction with AP and ROC-AUC, and destination node retrieval with Hits@K under both transductive and inductive settings. All reported results are averaged over five runs with different random seeds, and the standard deviation is reported. More details on dataset statistics, evaluation protocols, and implementation settings are provided in the Appendix~\ref{app:detail_setting}.

\subsection{Comparative Study}
\label{subsec:comparative_study}

We report the main comparative results on temporal link prediction and destination node retrieval. Table~\ref{tab:comp_result_ap} presents the AP results for temporal link prediction under transductive and inductive settings. Table~\ref{tab:comp_result_hits3} shows Hits@3 results for destination node retrieval. More complete results are provided in the Appendix~\ref{app:additional_results}.

\begin{table}[htbp]
    \centering
    \caption{AP (\%) for dynamic link prediction under transductive (\textit{tr.}) and inductive (\textit{in.}) settings. OOM means out of memory. The best and the second best results are marked as \textbf{bold} and \underline{underlined}, respectively.}
    \label{tab:comp_result_ap}
    \resizebox{\linewidth}{!}{
    \begin{tabular}{lcccccccccc}
    \toprule
     & \textbf{Datasets} & \textbf{JODIE} & \textbf{DyRep} & \textbf{TGAT} & \textbf{CAWN} & \textbf{TCL} & \textbf{GraphMixer} & \textbf{DyGFormer} & \textbf{MoMent} & \textbf{PRISM} \\
    \midrule
    \multirow{8}{*}{\textit{tr.}}
    & Enron
     & 95.53 $\pm$ 0.51
     & 90.66 $\pm$ 0.76
     & 96.68 $\pm$ 0.26
     & 97.56 $\pm$ 0.08
     & 96.03 $\pm$ 0.18
     & 95.59 $\pm$ 0.27
     & \underline{98.04 $\pm$ 0.15}
     & 96.73 $\pm$ 0.04
     & \textbf{98.19 $\pm$ 0.09} \\
     & ICEWS1819
     & 97.52 $\pm$ 0.37
     & 96.76 $\pm$ 0.26
     & 99.08 $\pm$ 0.32
     & 98.86 $\pm$ 0.25
     & \underline{99.27 $\pm$ 0.12}
     & 98.71 $\pm$ 0.34
     & 99.01 $\pm$ 0.18
     & 99.00 $\pm$ 0.00
     & \textbf{99.30 $\pm$ 0.02} \\
     & Googlemap CT
     & OOM
     & OOM
     & \textbf{90.02 $\pm$ 0.19}
     & 87.21 $\pm$ 0.27
     & 83.35 $\pm$ 0.18
     & 80.72 $\pm$ 0.10
     & 81.83 $\pm$ 0.38
     & 80.89 $\pm$ 0.04
     & \underline{87.89 $\pm$ 0.14} \\
     & GDELT
     & 94.66 $\pm$ 0.32
     & 94.16 $\pm$ 0.17
     & 95.72 $\pm$ 0.29
     & 95.82 $\pm$ 0.53
     & 96.01 $\pm$ 0.11
     & 95.23 $\pm$ 0.20
     & \underline{96.53 $\pm$ 0.03}
     & 96.01 $\pm$ 0.03
     & \textbf{96.99 $\pm$ 0.02} \\
     & Stack elec
     & OOM
     & OOM
     & 96.46 $\pm$ 0.05
     & 95.29 $\pm$ 0.23
     & 94.41 $\pm$ 0.79
     & 95.91 $\pm$ 0.09
     & \textbf{98.19 $\pm$ 0.10}
     & 95.21 $\pm$ 0.01
     & \underline{96.61 $\pm$ 0.13} \\
     & Stack ubuntu
     & OOM
     & OOM
     & 93.52 $\pm$ 0.12
     & 94.20 $\pm$ 0.05
     & 94.16 $\pm$ 0.23
     & 94.16 $\pm$ 0.47
     & 94.31 $\pm$ 0.08
     & \underline{95.18 $\pm$ 0.02}
     & \textbf{95.98 $\pm$ 0.15} \\
     & Amazon movies
     & OOM
     & OOM
     & 90.65 $\pm$ 0.16
     & 88.85 $\pm$ 0.07
     & 90.51 $\pm$ 0.10
     & 89.06 $\pm$ 0.06
     & \underline{90.97 $\pm$ 0.03}
     & 89.84 $\pm$ 0.04
     & \textbf{91.11 $\pm$ 0.17} \\
     & Yelp
     & OOM
     & OOM
     & 94.57 $\pm$ 0.25
     & 93.23 $\pm$ 0.27
     & \underline{95.11 $\pm$ 0.13}
     & 88.83 $\pm$ 0.26
     & 93.91 $\pm$ 0.11
     & 90.69 $\pm$ 0.07
     & \textbf{95.75 $\pm$ 0.07} \\
     \midrule
     & \textbf{Avg. Rank}
     & 8.31 & 8.69 & 3.88 & 4.75 & 4.25 & 6.38 & \underline{2.88} & 4.62 & \textbf{1.25} \\
     \midrule
    \multirow{8}{*}{\textit{in.}}
     & Enron
     & 87.61 $\pm$ 0.23
     & 77.34 $\pm$ 0.44
     & 85.89 $\pm$ 0.31
     & 92.23 $\pm$ 0.11
     & 85.60 $\pm$ 0.24
     & 83.28 $\pm$ 0.34
     & \underline{94.09 $\pm$ 0.25}
     & 88.74 $\pm$ 0.10
     & \textbf{94.13 $\pm$ 0.35} \\
     & ICEWS1819
     & 93.33 $\pm$ 0.26
     & 91.34 $\pm$ 0.41
     & 97.16 $\pm$ 0.33
     & 96.31 $\pm$ 0.34
     & \underline{97.89 $\pm$ 0.22}
     & 96.25 $\pm$ 0.30
     & 96.88 $\pm$ 0.18
     & 96.63 $\pm$ 0.03
     & \textbf{97.94 $\pm$ 0.05} \\
     & Googlemap CT
     & OOM
     & OOM
     & \textbf{87.50 $\pm$ 0.15}
     & 80.12 $\pm$ 0.21
     & 79.36 $\pm$ 0.09
     & 76.33 $\pm$ 0.13
     & 77.35 $\pm$ 0.31
     & 75.96 $\pm$ 0.05
     & \underline{85.25 $\pm$ 0.07} \\
     & GDELT
     & 90.19 $\pm$ 0.23
     & 89.25 $\pm$ 0.48
     & 90.23 $\pm$ 0.10
     & 89.86 $\pm$ 0.77
     & \underline{91.51 $\pm$ 0.45}
     & 89.25 $\pm$ 0.48
     & \textbf{92.63 $\pm$ 0.09}
     & 86.36 $\pm$ 0.07
     & 90.62 $\pm$ 0.03 \\
     & Stack elec
     & OOM
     & OOM
     & 83.91 $\pm$ 0.36
     & 79.21 $\pm$ 0.86
     & 75.99 $\pm$ 0.41
     & 81.42 $\pm$ 0.21
     & \textbf{88.01 $\pm$ 0.43}
     & 79.25 $\pm$ 0.06
     & \underline{84.61 $\pm$ 0.58} \\
     & Stack ubuntu
     & OOM
     & OOM
     & 76.64 $\pm$ 0.15
     & 78.82 $\pm$ 0.15
     & 77.77 $\pm$ 0.15
     & 78.70 $\pm$ 0.15
     & 78.32 $\pm$ 0.15
     & \underline{80.97 $\pm$ 0.08}
     & \textbf{83.89 $\pm$ 0.68} \\
     & Amazon movies
     & OOM
     & OOM
     & 87.60 $\pm$ 0.10
     & 85.08 $\pm$ 0.06
     & 87.38 $\pm$ 0.05
     & 85.17 $\pm$ 0.07
     & \underline{87.80 $\pm$ 0.06}
     & 86.07 $\pm$ 0.08
     & \textbf{88.01 $\pm$ 0.23} \\
     & Yelp
     & OOM
     & OOM
     & 91.74 $\pm$ 0.10
     & 90.10 $\pm$ 0.09
     & \underline{92.49 $\pm$ 0.13}
     & 84.94 $\pm$ 0.08
     & 90.92 $\pm$ 0.06
     & 87.16 $\pm$ 0.08
     & \textbf{93.41 $\pm$ 0.08} \\
     \midrule
     & \textbf{Avg. Rank}
     & 7.56 & 8.50 & 3.75 & 4.88 & 4.25 & 6.19 & \underline{3.00} & 5.38 & \textbf{1.50} \\
     \bottomrule
    \end{tabular}}
\end{table}

\paragraph{PRISM achieves the best overall performance across tasks and evaluation settings.}
As shown in Table~\ref{tab:comp_result_ap} and Table~\ref{tab:comp_result_hits3}, PRISM achieves the best average rank under both transductive and inductive settings, and remains consistently competitive across temporal link prediction and destination node retrieval. These results indicate that the advantage of PRISM comes from learning generally effective node states for DyTAG representation learning. Compared with TGNN baselines that mainly rely on historical interaction structures, PRISM further exploits textual semantics as stable node-level priors and adaptively refines them with behavioral evidence. Compared with MoMent, which follows a carrier-level text--structure--time partition and performs one-shot fusion, PRISM models a more intrinsic semantic--behavioral relationship and progressively transforms semantic priors into behavior-conditioned posterior states. The consistent gains across AP, AUC-ROC, and Hits@K suggest that PRISM learns generally effective DyTAG node states.

\paragraph{The dataset-level results reveal when semantic--behavioral posterior refinement is most beneficial.}
PRISM shows clear advantages on datasets where node semantics and interaction behaviors are tightly coupled. On Yelp and Amazon movies, textual descriptions and interaction histories jointly reflect user preferences and item characteristics, where PRISM achieves strong performance in both link prediction and retrieval. On Stack ubuntu, PRISM performs particularly well under the inductive setting, suggesting that semantic priors are useful when nodes have sparse or newly emerging histories. The results on Googlemap CT further show a nuanced pattern: although TGAT remains stronger for link prediction, PRISM substantially improves inductive retrieval, indicating that semantic priors can provide discriminative initialization for ranking history-sparse destinations. Meanwhile, PRISM is not uniformly dominant on all datasets. On Enron and GDELT, DyGFormer remains highly competitive in some retrieval settings, implying that datasets dominated by strong temporal regularities or event-driven interaction patterns may still benefit from powerful sequence-based temporal encoders. Nevertheless, PRISM maintains the best overall ranking across tasks and settings, demonstrating the robustness and transferability of semantic--behavioral posterior refinement for DyTAG learning.

\begin{table}[htbp]
    \centering
    \caption{Hits@3 (\%) for destination node retrieval under transductive (\textit{tr.}) and inductive (\textit{in.}) settings. The best and the second best results are marked as \textbf{bold} and \underline{underlined}, respectively.}
    \label{tab:comp_result_hits3}
    \resizebox{\linewidth}{!}{
    \begin{tabular}{lcccccccccc}
    \toprule
     & \textbf{Datasets} & \textbf{JODIE} & \textbf{DyRep} & \textbf{TGAT} & \textbf{CAWN} & \textbf{TCL} & \textbf{GraphMixer} & \textbf{DyGFormer} & \textbf{MoMent} & \textbf{PRISM} \\
    \midrule
    \multirow{6}{*}{\textit{tr.}}
     & Enron
     & 76.12 $\pm$ 3.11
     & OOM
     & 75.83 $\pm$ 0.82
     & 77.11 $\pm$ 0.64
     & 70.08 $\pm$ 0.81
     & 67.54 $\pm$ 0.27
     & \textbf{88.77 $\pm$ 0.51}
     & 74.86 $\pm$ 0.16
     & \underline{85.97 $\pm$ 0.48} \\
     & ICEWS1819
     & 85.32 $\pm$ 0.12
     & 84.15 $\pm$ 0.61
     & 91.88 $\pm$ 0.34
     & 89.51 $\pm$ 0.78
     & \underline{93.56 $\pm$ 1.06}
     & 92.31 $\pm$ 0.13
     & 91.75 $\pm$ 0.06
     & 91.07 $\pm$ 0.02
     & \textbf{93.61 $\pm$ 0.14} \\
     & Googlemap CT
     & OOM
     & OOM
     & \textbf{43.25 $\pm$ 0.23}
     & 27.71 $\pm$ 0.18
     & 29.66 $\pm$ 0.31
     & 25.14 $\pm$ 0.28
     & 27.05 $\pm$ 0.21
     & 25.58 $\pm$ 0.09
     & \underline{39.65 $\pm$ 0.10} \\
     & GDELT
     & 57.52 $\pm$ 0.35
     & 57.31 $\pm$ 0.28
     & 66.21 $\pm$ 0.19
     & 66.31 $\pm$ 0.17
     & 67.56 $\pm$ 0.29
     & 63.51 $\pm$ 0.30
     & \underline{70.89 $\pm$ 0.10}
     & 67.59 $\pm$ 0.24
     & \textbf{73.89 $\pm$ 0.06} \\
     & Amazon movies
     & OOM
     & OOM
     & 48.55 $\pm$ 0.13
     & 41.22 $\pm$ 0.65
     & 48.00 $\pm$ 0.11
     & 44.38 $\pm$ 0.13
     & \underline{49.90 $\pm$ 0.08}
     & 45.65 $\pm$ 1.33
     & \textbf{50.19 $\pm$ 0.50} \\
     & Yelp
     & OOM
     & OOM
     & \underline{59.68 $\pm$ 0.43}
     & 53.93 $\pm$ 0.24
     & 29.86 $\pm$ 0.84
     & 40.17 $\pm$ 0.42
     & 55.49 $\pm$ 0.54
     & 46.58 $\pm$ 0.01
     & \textbf{66.39 $\pm$ 0.46} \\
     \midrule
     & \textbf{Avg. Rank}
     & 7.67 & 8.33 & 3.50 & 5.00 & 4.50 & 6.00 & \underline{3.00} & 5.67 & \textbf{1.33} \\
     \midrule
    \multirow{6}{*}{\textit{in.}}
     & Enron
     & 52.44 $\pm$ 0.16
     & OOM
     & 50.45 $\pm$ 0.25
     & 59.62 $\pm$ 0.27
     & 39.28 $\pm$ 0.35
     & 39.16 $\pm$ 0.11
     & \textbf{74.74 $\pm$ 0.26}
     & 45.16 $\pm$ 0.45
     & \underline{70.36 $\pm$ 0.20} \\
     & ICEWS1819
     & 71.40 $\pm$ 0.86
     & 70.10 $\pm$ 0.32
     & 78.35 $\pm$ 0.29
     & \underline{81.20 $\pm$ 0.66}
     & 80.26 $\pm$ 0.91
     & 80.01 $\pm$ 0.88
     & 80.17 $\pm$ 0.71
     & 75.74 $\pm$ 0.22
     & \textbf{84.13 $\pm$ 0.54} \\
     & Googlemap CT
     & OOM
     & OOM
     & \underline{15.45 $\pm$ 0.17}
     & 8.28 $\pm$ 0.21
     & 6.21 $\pm$ 0.08
     & 4.65 $\pm$ 0.04
     & 4.48 $\pm$ 0.07
     & 20.46 $\pm$ 0.06
     & \textbf{33.77 $\pm$ 0.22} \\
     & GDELT
     & \underline{52.94 $\pm$ 0.28}
     & 51.79 $\pm$ 0.68
     & 48.18 $\pm$ 0.55
     & 46.67 $\pm$ 0.61
     & 50.90 $\pm$ 0.24
     & 47.55 $\pm$ 0.81
     & \textbf{56.66 $\pm$ 0.26}
     & 35.58 $\pm$ 0.24
     & 47.79 $\pm$ 0.19 \\
     & Amazon movies
     & OOM
     & OOM
     & 41.65 $\pm$ 0.61
     & 34.41 $\pm$ 0.51
     & 41.57 $\pm$ 0.23
     & 37.64 $\pm$ 0.11
     & \underline{43.19 $\pm$ 0.11}
     & 38.42 $\pm$ 1.41
     & \textbf{43.23 $\pm$ 0.57} \\
     & Yelp
     & OOM
     & OOM
     & \underline{51.20 $\pm$ 2.06}
     & 45.53 $\pm$ 1.20
     & 25.94 $\pm$ 1.68
     & 34.13 $\pm$ 0.60
     & 47.51 $\pm$ 0.64
     & 39.48 $\pm$ 0.01
     & \textbf{57.63 $\pm$ 0.46} \\
     \midrule
     & \textbf{Avg. Rank}
     & 6.67 & 7.33 & 3.83 & 4.50 & 4.83 & 6.17 & \underline{2.83} & 6.83 & \textbf{2.00} \\
     \bottomrule
    \end{tabular}}
\end{table}

\subsection{Ablation Studies}
\label{sec:ablation_study}

We conduct ablation studies to examine the effectiveness of the semantic--behavioral modality decomposition, the iterative posterior refinement process, and the auxiliary objectives. Specifically, \textit{w/o Semantic} removes the semantic prior and relies only on behavioral evidence, while \textit{w/o Behavior} removes historical behavioral evidence and relies only on semantic priors. For auxiliary objectives, \textit{w/o} $\mathcal{L}_{\mathrm{margin}}$, \textit{w/o} $\mathcal{L}_{\mathrm{step}}$, and \textit{w/o} $\mathcal{L}_{\mathrm{recon}}$ remove the semantic trust-region regularizer, the trajectory smoothness regularizer, and the behavioral reconstruction regularizer, respectively.

\begin{table}[htbp]
    \centering
    \caption{AP (\%) for modality ablation under transductive (\textit{tr.}) and inductive (\textit{in.}) settings. $\downarrow$ denotes the performance drop compared with the original PRISM.}
    \label{tab:ablation_modality}
    \resizebox{\linewidth}{!}{
    \begin{tabular}{lcccccc}
    \toprule
    \multirow{2}{*}{\textbf{Variant}} & \multicolumn{3}{c}{\textit{tr.}} & \multicolumn{3}{c}{\textit{in.}} \\
    \cmidrule(lr){2-4} \cmidrule(lr){5-7}
     & \textbf{Enron} & \textbf{Stack ubuntu} & \textbf{Googlemap CT} & \textbf{Enron} & \textbf{Stack ubuntu} & \textbf{Googlemap CT} \\
    \midrule
    PRISM & 98.19 & 95.98 & 87.89 & 94.13 & 83.89 & 85.25 \\
    w/o Semantic & 96.91 ($\downarrow$1.28) & 94.28 ($\downarrow$1.70) & 86.84 ($\downarrow$1.05) & 91.96 ($\downarrow$2.17) & 82.10 ($\downarrow$1.79) & 83.48 ($\downarrow$1.77) \\
    w/o Behavior & 86.58 ($\downarrow$11.61) & 65.21 ($\downarrow$30.77) & 73.69 ($\downarrow$14.20) & 72.35 ($\downarrow$21.78) & 59.53 ($\downarrow$24.36) & 68.43 ($\downarrow$16.82) \\
    \bottomrule
    \end{tabular}}
\end{table}

\paragraph{Both semantic priors and behavioral evidence are important to PRISM.}
As shown in Table~\ref{tab:ablation_modality}, removing either modality leads to consistent performance degradation, validating the effectiveness of the proposed semantic--behavioral decomposition. Removing the semantic modality weakens the model across all datasets, with more evident drops under the inductive setting. This observation suggests that textual semantics provide useful node-level priors, especially when nodes have sparse or newly emerging interaction histories. Removing the behavioral modality causes much larger performance drops, indicating that historical interactions remain essential for modeling time-dependent contexts and dynamic preferences. These results show that the two modalities play complementary roles: semantic priors provide stable identity-level information, while behavioral evidence supplies dynamic observations for posterior refinement.

\begin{wrapfigure}{R}{0.6\linewidth}
    \centering
    \begin{subfigure}[b]{0.48\linewidth}
        \centering
        \includegraphics[width=\linewidth]{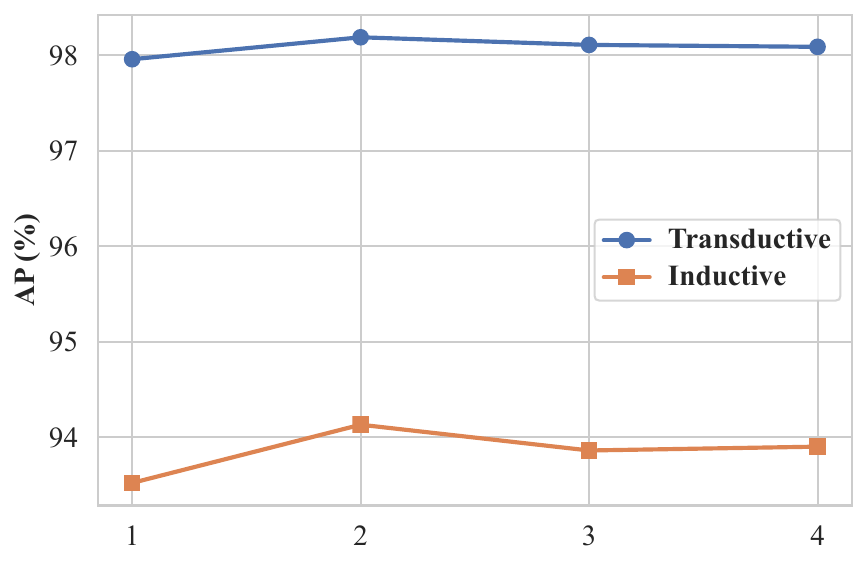}
        \caption{Enron}
    \end{subfigure}
    \hfill
    \begin{subfigure}[b]{0.48\linewidth}
        \centering
        \includegraphics[width=\linewidth]{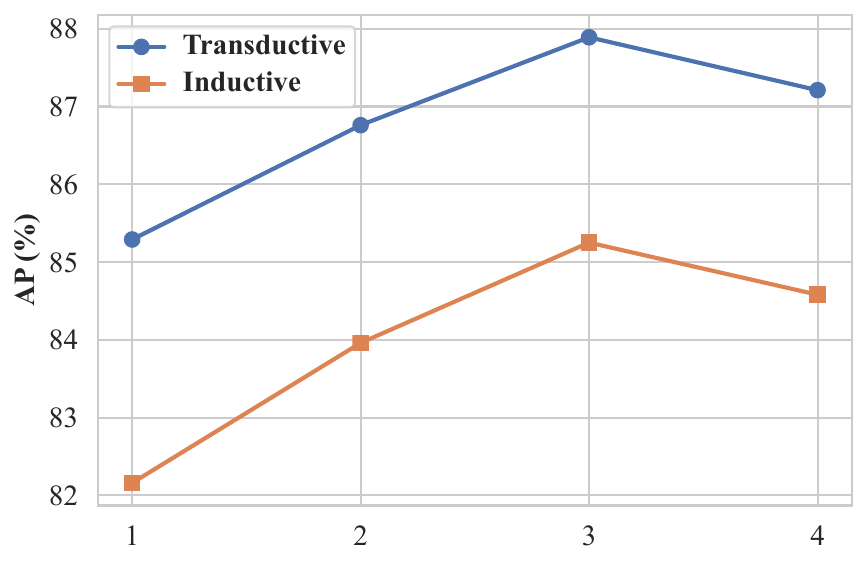}
        \caption{Googlemap CT}
    \end{subfigure}
    \caption{AP (\%) for refinement-step ablation}
    \label{fig:ablation_step}
\end{wrapfigure}

\paragraph{Iterative refinement benefits from a moderate number of steps.}
Figure~\ref{fig:ablation_step} studies the effect of the number of refinement steps. Compared with one-step refinement, using multiple refinement steps generally improves performance, demonstrating that posterior refinement is more effective when semantic priors are updated progressively rather than fused with behavioral evidence in a single pass. On Googlemap CT, the performance increases from 85.29 at $K=1$ to 87.89 at $K=3$, showing that additional refinement steps help the model better incorporate behavioral evidence in a more challenging retrieval-oriented scenario. On Enron, the best result is obtained at $K=2$, while further increasing $K$ brings only marginal changes. These results indicate that a small number of refinement steps is sufficient to capture the prior-to-posterior transition, whereas overly deep refinement may introduce redundant corrections or overfit noisy behavioral signals.

\begin{table}[htbp]
    \centering
    \caption{AUC-ROC (\%) for auxiliary-objective ablation under transductive (\textit{tr.}) and inductive (\textit{in.}) settings. $\downarrow$ denotes the performance drop compared with PRISM.}
    \label{tab:ablation_loss}
    \resizebox{\linewidth}{!}{
    \begin{tabular}{lcccccc}
    \toprule
    \multirow{2}{*}{\textbf{Variant}} & \multicolumn{3}{c}{\textit{tr.}} & \multicolumn{3}{c}{\textit{in.}} \\
    \cmidrule(lr){2-4} \cmidrule(lr){5-7}
     & \textbf{Enron} & \textbf{Stack elec} & \textbf{Googlemap CT} & \textbf{Enron} & \textbf{Stack elec} & \textbf{Googlemap CT} \\
    \midrule
    PRISM & 98.13 & 97.22 & 88.00 & 93.17 & 84.78 & 85.21 \\
    w/o $\mathcal{L}_{\mathrm{margin}}$ & 98.04 ($\downarrow$0.09) & 96.40 ($\downarrow$0.82) & 86.68 ($\downarrow$1.32) & 93.00 ($\downarrow$0.17) & 80.26 ($\downarrow$4.52) & 83.46 ($\downarrow$1.75) \\
    w/o $\mathcal{L}_{\mathrm{step}}$ & 97.98 ($\downarrow$0.15) & 96.49 ($\downarrow$0.73) & 86.16 ($\downarrow$1.84) & 92.23 ($\downarrow$0.94) & 80.71 ($\downarrow$4.07) & 83.08 ($\downarrow$2.13) \\
    w/o $\mathcal{L}_{\mathrm{recon}}$ & 97.86 ($\downarrow$0.27) & 96.25 ($\downarrow$0.97) & 86.55 ($\downarrow$1.45) & 92.34 ($\downarrow$0.83) & 79.28 ($\downarrow$5.50) & 83.61 ($\downarrow$1.60) \\
    \bottomrule
    \end{tabular}}
\end{table}

\paragraph{The auxiliary objectives stabilize posterior refinement.}
Table~\ref{tab:ablation_loss} reports the effect of removing each auxiliary objective. The performance consistently drops when any of the three objectives is removed, showing that they regularize the refinement process from complementary perspectives. Removing $\mathcal{L}_{\mathrm{margin}}$ weakens the semantic trust-region constraint, making the posterior states more likely to drift away from their semantic priors. Removing $\mathcal{L}_{\mathrm{step}}$ degrades performance by reducing the smoothness of the refinement trajectory, especially on datasets where behavioral evidence may be noisy or sparse. Removing $\mathcal{L}_{\mathrm{recon}}$ causes the largest drop in several inductive settings, suggesting that preserving behavior-relevant information in the final posterior state is particularly important for generalization. Overall, these results confirm that the auxiliary objectives help PRISM learn posterior states that are behavior-aware, semantically anchored, and smoothly refined.

\section{Conclusion}
\label{sec:conclusion}

In this paper, we proposed PRISM for DyTAG representation learning, which models DyTAGs from a semantic--behavioral perspective and performs iterative posterior refinement for node representation learning. Extensive experiments on DTGB benchmark datasets demonstrate that PRISM achieves strong performance on temporal link prediction and destination node retrieval under both transductive and inductive settings, while ablation studies verify the effectiveness of its key components. Currently, PRISM mainly adapts ideas from generative modeling to discriminative DyTAG tasks. Future work will extend this framework toward generative DyTAG modeling scenarios, such as those in the GDGB benchmark~\cite{peng2026gdgb}.

\bibliography{references}

\appendix

\section{Notations}
\label{app:notations}

In this section, we summarize the important notations used in this paper, as detailed in Table~\ref{tab:notation}.

\begin{table}[htbp]
    \centering
    \caption{Notations and descriptions.}
    \label{tab:notation}
    \begin{tabular}{l p{0.8\linewidth}} 
    \toprule
    \textbf{Notation} & \textbf{Description} \\
    \midrule
    $\mathcal{G}$ & A dynamic text-attributed graph (DyTAG). \\
    $\mathcal{V}$ & The node set of the DyTAG. \\
    $\mathcal{E}$ & The chronological interaction set of the DyTAG. \\
    $\mathcal{G}_{<t}$ & The historical DyTAG consisting of all interactions before timestamp $t$. \\
    $\mathcal{H}_{u}^{t}$ & The chronological historical interaction sequence of node $u$ before timestamp $t$. \\
    \midrule
    $\mathbf{s}_u$ & The semantic prior representation of node $u$ encoded from its textual attribute. \\
    $\mathbf{h}_{u,i}^{t}$ & The behavioral token of the $i$-th historical interaction of node $u$ before timestamp $t$. \\
    $\mathbf{H}_{u}^{t}$ & The behavioral token sequence of node $u$ before timestamp $t$. \\
    $\mathbf{B}_{u}^{t}$ & The contextualized behavioral token sequence of node $u$ after behavioral encoding. \\
    $\mathbf{b}_{u}^{t}$ & The pooled behavioral representation of node $u$ before timestamp $t$. \\
    \midrule
    $\mathbf{z}_{u}^{(k,t)}$ & The intermediate posterior state of node $u$ at timestamp $t$ after the $k$-th refinement step. \\
    $\mathbf{z}_{u}^{t}$ & The final behavior-conditioned posterior representation of node $u$ at timestamp $t$. \\
    $\mathbf{c}_{u}^{(k,t)}$ & The behavioral context retrieved by cross-modal attention at the $k$-th refinement step. \\
    $\Delta \mathbf{z}_{u}^{(k,t)}$ & The refinement direction estimated at the $k$-th refinement step. \\
    \midrule
    $\hat{y}_{uv}^{t}$ & The predicted interaction probability between nodes $u$ and $v$ at timestamp $t$. \\
    $\mathcal{B}$ & A mini-batch of observed temporal interactions. \\
    $\mathcal{V}_{h}$ & The set of node-time instances with non-empty histories in a mini-batch. \\
    \midrule
    $\mathcal{L}_{\mathrm{task}}$ & The supervised task loss for downstream prediction. \\
    $\mathcal{L}_{\mathrm{recon}}$ & The behavioral reconstruction objective. \\
    $\mathcal{L}_{\mathrm{margin}}$ & The semantic trust-region objective. \\
    $\mathcal{L}_{\mathrm{step}}$ & The trajectory smoothness objective. \\
    $\mathcal{L}$ & The overall training objective of PRISM. \\
    \bottomrule
    \end{tabular}
\end{table}

\section{Detailed Experimental Settings}
\label{app:detail_setting}

\subsection{Evaluation Tasks and Metrics}
\label{app:evaluation_tasks}

Following the DTGB benchmark~\cite{zhang2024dtgb}, we evaluate all methods on two DyTAG representation learning tasks: temporal link prediction and destination node retrieval. For temporal link prediction, given a source node, a destination node, and a timestamp, the model predicts whether the candidate interaction will occur at the given time. We report average precision (AP) and ROC-AUC as evaluation metrics. For destination node retrieval, the model ranks the ground-truth destination node among a set of candidate nodes for each source-time query. We report Hits@K, which measures the proportion of test instances where the ground-truth destination node is ranked within the top-$K$ candidates.

We evaluate both tasks under transductive and inductive settings. In the transductive setting, test interactions are formed by nodes observed during training, while the interactions themselves occur after the training period. In the inductive setting, test interactions involve nodes that are not observed in the training interactions, which evaluates the model's ability to generalize to newly emerging nodes. All evaluations follow the chronological split protocol of DTGB to avoid temporal information leakage.

\subsection{Implementation Details}
\label{app:implementation_details}

We adopt the experimental protocol provided by the DTGB benchmark~\cite{zhang2024dtgb}. For each dataset, temporal interactions are split into training, validation, and test sets with a ratio of 0.7:0.15:0.15 according to chronological order. All baseline models are trained and evaluated under the same data splits and evaluation protocol. For baseline methods, we use the recommended hyperparameter settings from the DTGB benchmark whenever applicable. All experiments were conducted on NVIDIA RTX 5090 GPUs with 32GB memory. Depending on the dataset scale, each run takes approximately 3--27 hours for PRISM.

For PRISM, node textual attributes and interaction textual attributes are encoded with a lightweight pre-trained language model, instantiated as \texttt{bert-base-uncased}~\cite{devlin2019bert}, and then projected into the model latent space. The behavioral history length is set to $L$ for sampling recent historical interactions before the prediction time. PRISM is optimized with the Adam optimizer~\cite{kingma2014adam}. Following the common DTGB setup, we use binary cross-entropy as the supervised objective for temporal link prediction, where each observed interaction is paired with a randomly sampled negative destination node. The final objective further includes the auxiliary posterior refinement objectives described in Section~\ref{subsec:training_objective}. Model selection is performed according to validation performance, and the best checkpoint is used for test evaluation.

\subsection{Datasets}
\label{app:datasets}

We evaluate PRISM on eight datasets from the DTGB benchmark~\cite{zhang2024dtgb}: Enron, ICEWS1819, Googlemap CT, GDELT, Stack elec, Stack ubuntu, Amazon movies, and Yelp. As shown in Table~\ref{tab:dataset_stat}, these datasets cover diverse DyTAG scenarios, including email communication networks, temporal knowledge graphs, online question-answering communities, e-commerce review graphs, and local service review graphs. Each dataset contains a chronological interaction list, node textual attributes, and edge or interaction textual attributes.

\begin{table}[htbp]
    \centering
    \caption{Statistics of datasets.}
    \label{tab:dataset_stat}
    \begin{tabular}{l c c c c c}
    \toprule
    \textbf{Dataset} & \textbf{Nodes} & \textbf{Edges} & \textbf{Timestamps} & \textbf{Domain} & \textbf{Bipartite} \\
    \midrule
    Enron            & 42,711         & 797,907        & 1,006               & E-mail           & $\times$ \\
    GDELT            & 6,786          & 1,339,245      & 2,591               & Knowledge graph  & $\times$ \\
    ICEWS1819        & 31,796         & 1,100,071      & 730                 & Knowledge graph  & $\times$ \\
    Stack elec       & 397,702        & 1,262,225      & 5,224               & Multi-round dialogue & $\checkmark$ \\
    Stack ubuntu     & 674,248        & 1,497,006      & 4,972               & Multi-round dialogue & $\checkmark$ \\
    Googlemap CT     & 111,168        & 1,380,623      & 55,521              & E-commerce       & $\checkmark$ \\
    Amazon movies    & 293,566        & 3,217,324      & 7,287               & E-commerce       & $\checkmark$ \\
    Yelp             & 2,138,242      & 6,990,189      & 6,036               & E-commerce       & $\checkmark$ \\
    \bottomrule
    \end{tabular}
\end{table}

\begin{itemize}[leftmargin=15pt, itemsep=1pt]
    \item \textbf{Enron} is derived from the email communications of employees at the Enron corporation. Nodes represent employees, while temporal edges represent emails between them. Node textual attributes describe employee information when available, and edge textual attributes correspond to email contents.
    
    \item \textbf{ICEWS1819} is constructed from the Integrated Crisis Early Warning System and contains political events from 2018 to 2019. Nodes represent political entities, and edges represent temporal political interactions with relation descriptions.
    
    \item \textbf{Googlemap CT} is extracted from the Connecticut subset of Google Local Data. It forms a bipartite review graph where nodes are users and businesses, and temporal edges correspond to user reviews. Business descriptions and review texts provide textual attributes.
    
    \item \textbf{GDELT} is derived from the Global Database of Events, Language, and Tone. Nodes represent political entities, while edges describe temporal political events or relationships between entities.
    
    \item \textbf{Stack elec} is constructed from Stack Exchange data related to electronic techniques. Nodes correspond to users and questions, while edges represent answers or comments with textual contents.
    
    \item \textbf{Stack ubuntu} is another Stack Exchange dataset focused on Ubuntu-related questions. Compared with Stack elec, its textual interactions often contain a mixture of natural language and code, making semantic modeling more challenging.
    
    \item \textbf{Amazon movies} is extracted from Amazon review data in the Movies and TV category. Nodes represent users and products, while temporal edges correspond to product reviews with rating categories and review texts.
    
    \item \textbf{Yelp} is derived from the Yelp Open Dataset. Nodes represent users and businesses, and temporal edges correspond to user reviews. Business profiles, user information, and review texts serve as textual attributes.
\end{itemize}

\subsection{Baselines}
\label{app:baselines}

We compare PRISM with representative temporal graph learning and DyTAG modeling baselines. The temporal graph baselines include memory-based methods, temporal attention methods, random-walk-based methods, and recent sequence-based dynamic graph models. We also include MoMent as a multimodal DyTAG learning baseline.

\begin{itemize}[leftmargin=15pt, itemsep=1pt]
    \item \textbf{JODIE}~\cite{kumar2019jodie} learns dynamic embedding trajectories for interacting entities with coupled recurrent neural networks and uses a projection operator to estimate future node states.
    
    \item \textbf{DyRep}~\cite{trivedi2019dyrep} models dynamic graph evolution with recurrent node state updates and temporal point processes, capturing both interaction dynamics and structural evolution.
    
    \item \textbf{TGAT}~\cite{xu2020tgat} introduces temporal graph attention with functional time encoding to aggregate temporal-topological neighborhoods for inductive dynamic graph representation learning.
    
    \item \textbf{CAWN}~\cite{wang2021cawn} captures temporal dependencies through causal anonymous walks and learns inductive node representations from sampled temporal walk patterns.
    
    \item \textbf{TCL}~\cite{wang2021tcl} adopts a two-stream Transformer architecture to encode temporal neighborhoods of target interaction nodes and uses contrastive learning for dynamic graph representation.
    
    \item \textbf{GraphMixer}~\cite{cong2023graphmixer} is a lightweight temporal graph model based on MLP mixing and neighbor mean-pooling, showing strong performance without recurrent or attention-heavy architectures.
    
    \item \textbf{DyGFormer}~\cite{yu2023dygformer} models historical first-hop interaction sequences with neighbor co-occurrence encoding and sequence patching, enabling effective long-history temporal graph modeling.
    
    \item \textbf{MoMent}~\cite{xu2026moment} is a multimodal DyTAG learning method that models textual, structural, and temporal modalities with modality-specific encoders and integrates them for dynamic graph representation learning.
\end{itemize}

\subsection{Training of PRISM}
\label{app:training_prism}

Given a positive temporal interaction $(u,v,t)$, PRISM first constructs the semantic prior representations and behavioral evidence for the source node $u$ and the destination node $v$ based only on the historical graph before timestamp $t$. The iterative posterior refinement module then produces the time-dependent posterior states $\mathbf{z}_u^t$ and $\mathbf{z}_v^t$. A prediction MLP takes the concatenation of the two posterior states as input and outputs the interaction probability.

For each positive interaction, we randomly sample a negative destination node $v^{-}$ to construct a negative interaction $(u,v^{-},t)$. The model is trained to assign higher scores to observed interactions than to negative samples through binary cross-entropy loss. The supervised task loss is combined with the auxiliary objectives for behavioral reconstruction, semantic trust-region regularization, and trajectory smoothness regularization. During evaluation, the model scores candidate destinations using the refined posterior states and ranks them according to the predicted interaction scores.

\section{Additional Experiment Results}
\label{app:additional_results}

This section provides additional experimental results for temporal link prediction and destination node retrieval. Table~\ref{tab:comp_result_auc} reports the AUC-ROC results for dynamic link prediction. Consistent with the AP results in the main text, PRISM achieves the best average rank under both transductive and inductive settings, indicating that the effectiveness of PRISM is stable across different link prediction metrics.

\begin{table}[htbp]
    \centering
    \caption{AUC-ROC (\%) for dynamic link prediction under transductive (\textit{tr.}) and inductive (\textit{in.}) settings. The best and the second best results are marked as \textbf{bold} and \underline{underlined}, respectively.}
    \label{tab:comp_result_auc}
    \resizebox{\linewidth}{!}{
    \begin{tabular}{lcccccccccc}
    \toprule
     & \textbf{Datasets} & \textbf{JODIE} & \textbf{DyRep} & \textbf{TGAT} & \textbf{CAWN} & \textbf{TCL} & \textbf{GraphMixer} & \textbf{DyGFormer} & \textbf{MoMent} & \textbf{PRISM} \\
    \midrule
    \multirow{8}{*}{\textit{tr.}}
     & Enron
     & 97.31 $\pm$ 0.52
     & 92.74 $\pm$ 0.26
     & 96.81 $\pm$ 0.26
     & 97.40 $\pm$ 0.07
     & 96.18 $\pm$ 0.25
     & 95.67 $\pm$ 0.13
     & \underline{97.79 $\pm$ 0.14}
     & 96.96 $\pm$ 0.04
     & \textbf{98.13 $\pm$ 0.09} \\
     & ICEWS1819
     & 97.41 $\pm$ 1.13
     & 96.32 $\pm$ 0.27
     & 99.04 $\pm$ 0.39
     & 98.57 $\pm$ 0.18
     & \underline{99.23 $\pm$ 0.12}
     & 98.63 $\pm$ 0.24
     & 98.88 $\pm$ 0.15
     & 98.97 $\pm$ 0.00
     & \textbf{99.24 $\pm$ 0.02} \\
     & Googlemap CT
     & OOM
     & OOM
     & \textbf{90.49 $\pm$ 0.71}
     & 86.87 $\pm$ 0.63
     & 83.48 $\pm$ 0.94
     & 80.95 $\pm$ 0.14
     & 82.07 $\pm$ 0.18
     & 81.40 $\pm$ 0.06
     & \underline{88.00 $\pm$ 0.23} \\
     & GDELT
     & 95.33 $\pm$ 0.20
     & 94.53 $\pm$ 0.18
     & 95.95 $\pm$ 0.33
     & 96.00 $\pm$ 0.61
     & 96.19 $\pm$ 0.08
     & 95.52 $\pm$ 0.18
     & \underline{96.62 $\pm$ 0.03}
     & 96.21 $\pm$ 0.03
     & \textbf{97.08 $\pm$ 0.02} \\
     & Stack elec
     & OOM
     & OOM
     & 97.09 $\pm$ 0.14
     & 96.31 $\pm$ 0.07
     & 95.78 $\pm$ 1.06
     & 96.73 $\pm$ 0.11
     & \textbf{97.98 $\pm$ 0.06}
     & 96.18 $\pm$ 0.02
     & \underline{97.22 $\pm$ 0.07} \\
     & Stack ubuntu
     & OOM
     & OOM
     & 94.90 $\pm$ 0.18
     & 95.67 $\pm$ 0.04
     & 95.16 $\pm$ 0.03
     & 94.94 $\pm$ 0.28
     & 95.26 $\pm$ 0.35
     & \underline{95.92 $\pm$ 0.01}
     & \textbf{96.49 $\pm$ 0.07} \\
     & Amazon movies
     & OOM
     & OOM
     & 90.64 $\pm$ 0.14
     & 89.36 $\pm$ 0.12
     & 90.50 $\pm$ 0.12
     & 88.94 $\pm$ 0.08
     & \underline{91.00 $\pm$ 0.06}
     & 89.73 $\pm$ 0.04
     & \textbf{91.12 $\pm$ 0.12} \\
     & Yelp
     & OOM
     & OOM
     & 94.87 $\pm$ 0.29
     & 93.49 $\pm$ 0.26
     & \underline{95.28 $\pm$ 0.18}
     & 89.27 $\pm$ 0.21
     & 94.07 $\pm$ 0.10
     & 90.88 $\pm$ 0.06
     & \textbf{95.97 $\pm$ 0.06} \\
     \midrule
     & \textbf{Avg. Rank}
     & 7.81 & 8.69 & 4.00 & 4.62 & 4.38 & 6.50 & \underline{3.12} & 4.62 & \textbf{1.25} \\
     \midrule
    \multirow{8}{*}{\textit{in.}}
     & Enron
     & 87.32 $\pm$ 0.37
     & 79.01 $\pm$ 0.47
     & 86.50 $\pm$ 0.32
     & 90.91 $\pm$ 0.14
     & 85.12 $\pm$ 0.62
     & 83.47 $\pm$ 0.39
     & \underline{93.16 $\pm$ 0.15}
     & 88.52 $\pm$ 0.12
     & \textbf{93.17 $\pm$ 0.51} \\
     & ICEWS1819
     & 92.85 $\pm$ 0.65
     & 90.30 $\pm$ 0.97
     & 97.06 $\pm$ 0.54
     & \underline{97.74 $\pm$ 0.39}
     & \textbf{97.78 $\pm$ 0.12}
     & 96.05 $\pm$ 0.25
     & 96.30 $\pm$ 0.27
     & 96.57 $\pm$ 0.02
     & \textbf{97.78 $\pm$ 0.06} \\
     & Googlemap CT
     & OOM
     & OOM
     & \textbf{87.91 $\pm$ 0.28}
     & 70.58 $\pm$ 0.47
     & 78.95 $\pm$ 0.46
     & 75.43 $\pm$ 0.18
     & 76.48 $\pm$ 0.52
     & 75.22 $\pm$ 0.06
     & \underline{85.21 $\pm$ 0.12} \\
     & GDELT
     & 89.21 $\pm$ 0.65
     & 89.17 $\pm$ 0.07
     & 90.12 $\pm$ 0.11
     & 88.99 $\pm$ 0.82
     & \underline{90.99 $\pm$ 0.22}
     & 89.42 $\pm$ 0.35
     & \textbf{92.06 $\pm$ 0.03}
     & 86.60 $\pm$ 0.11
     & 90.31 $\pm$ 0.05 \\
     & Stack elec
     & OOM
     & OOM
     & 84.23 $\pm$ 0.18
     & 79.63 $\pm$ 0.74
     & 76.89 $\pm$ 0.51
     & 82.32 $\pm$ 0.31
     & \textbf{86.07 $\pm$ 0.15}
     & 79.22 $\pm$ 0.05
     & \underline{84.78 $\pm$ 0.36} \\
     & Stack ubuntu
     & OOM
     & OOM
     & 76.55 $\pm$ 0.19
     & 78.71 $\pm$ 0.80
     & 77.17 $\pm$ 0.27
     & 78.70 $\pm$ 0.32
     & 77.73 $\pm$ 0.47
     & \underline{80.28 $\pm$ 0.07}
     & \textbf{83.29 $\pm$ 0.57} \\
     & Amazon movies
     & OOM
     & OOM
     & 87.06 $\pm$ 0.23
     & 84.92 $\pm$ 0.08
     & 86.77 $\pm$ 0.09
     & 84.18 $\pm$ 0.12
     & \underline{87.33 $\pm$ 0.05}
     & 85.14 $\pm$ 0.08
     & \textbf{87.50 $\pm$ 0.18} \\
     & Yelp
     & OOM
     & OOM
     & 91.73 $\pm$ 0.08
     & 89.95 $\pm$ 0.05
     & \underline{92.33 $\pm$ 0.11}
     & 84.52 $\pm$ 0.14
     & 90.67 $\pm$ 0.09
     & 86.59 $\pm$ 0.10
     & \textbf{93.29 $\pm$ 0.09} \\
     \midrule
     & \textbf{Avg. Rank}
     & 7.69 & 8.44 & 3.88 & 5.00 & 4.12 & 5.88 & \underline{3.12} & 5.38 & \textbf{1.50} \\
     \bottomrule
    \end{tabular}}
\end{table}

\begin{figure}[htbp]
    \centering
    \begin{subfigure}[b]{0.48\linewidth}
        \centering
        \includegraphics[width=\linewidth]{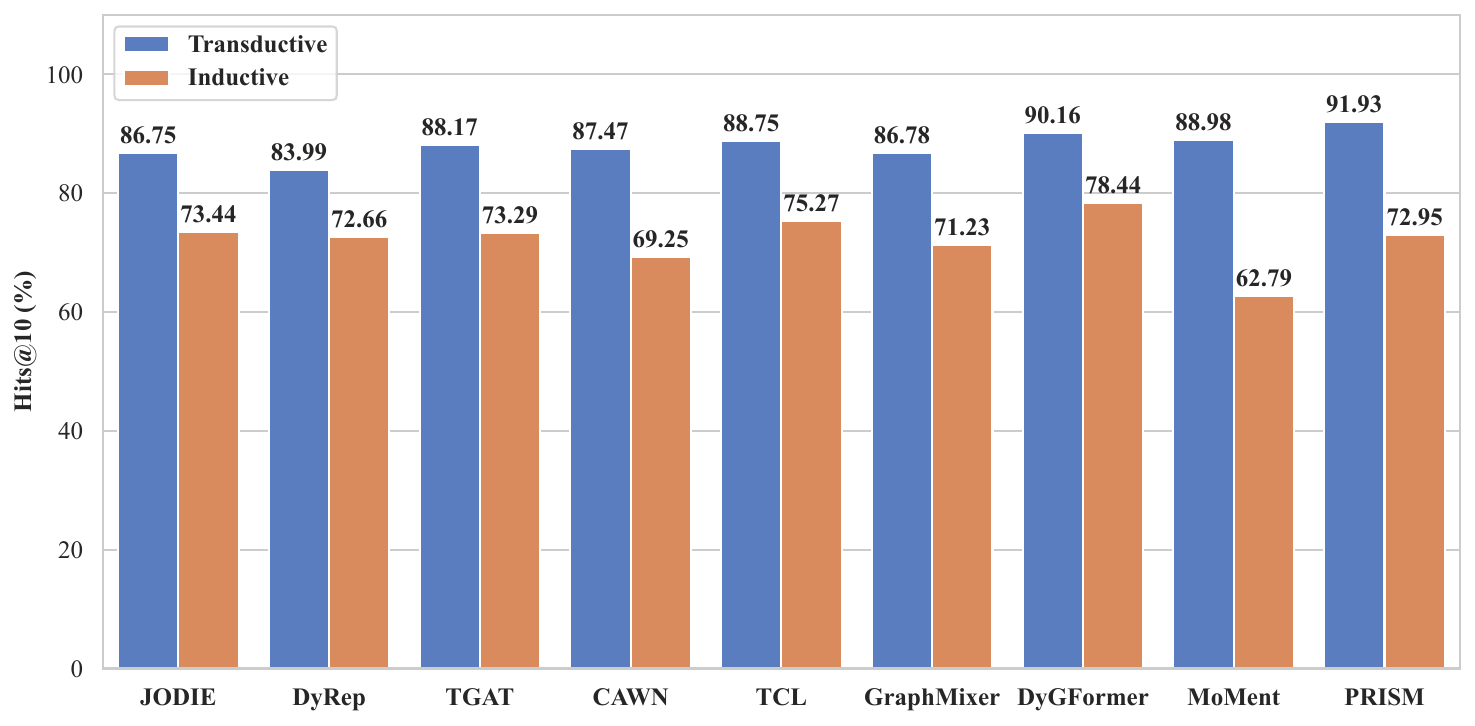}
        \caption{GDELT}
    \end{subfigure}
    \hfill
    \begin{subfigure}[b]{0.48\linewidth}
        \centering
        \includegraphics[width=\linewidth]{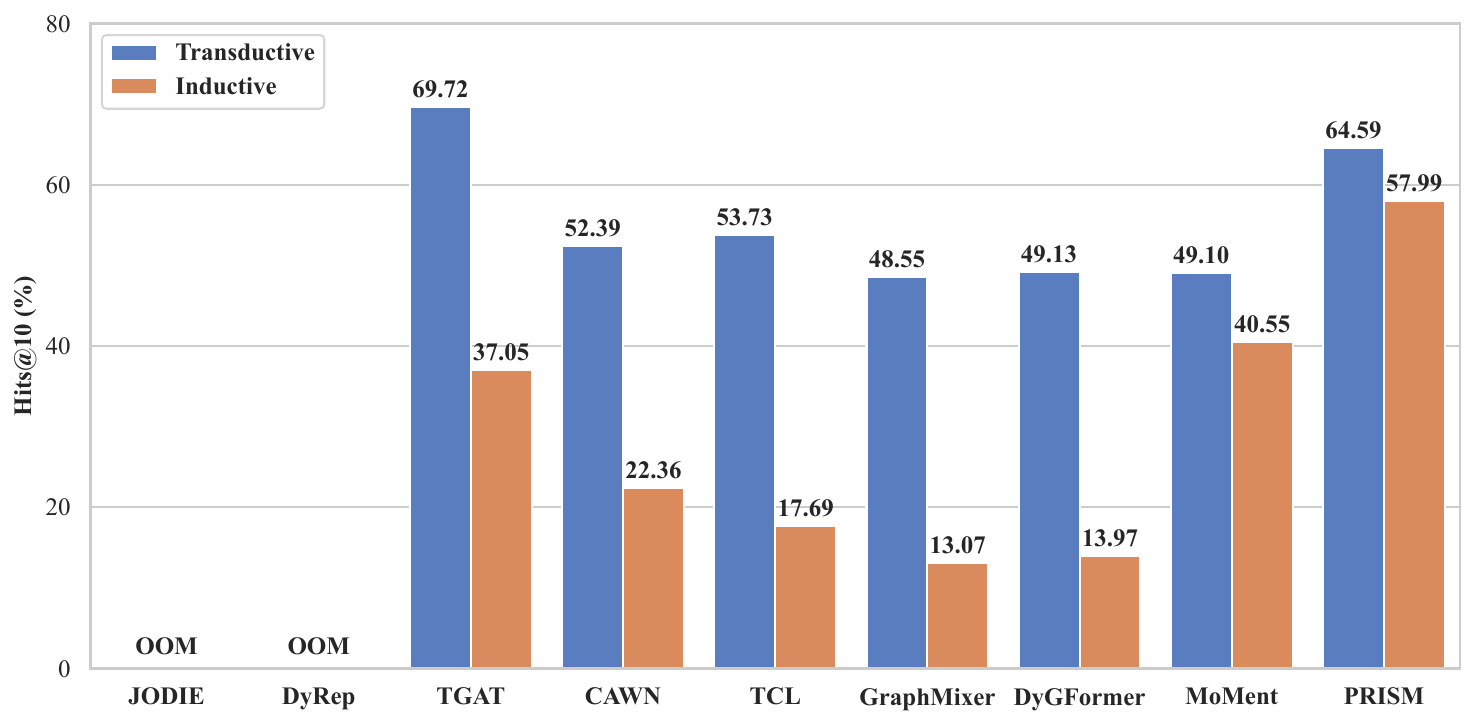}
        \caption{Googlemap CT}
    \end{subfigure}
    \\[2mm]
    \begin{subfigure}[b]{0.48\linewidth}
        \centering
        \includegraphics[width=\linewidth]{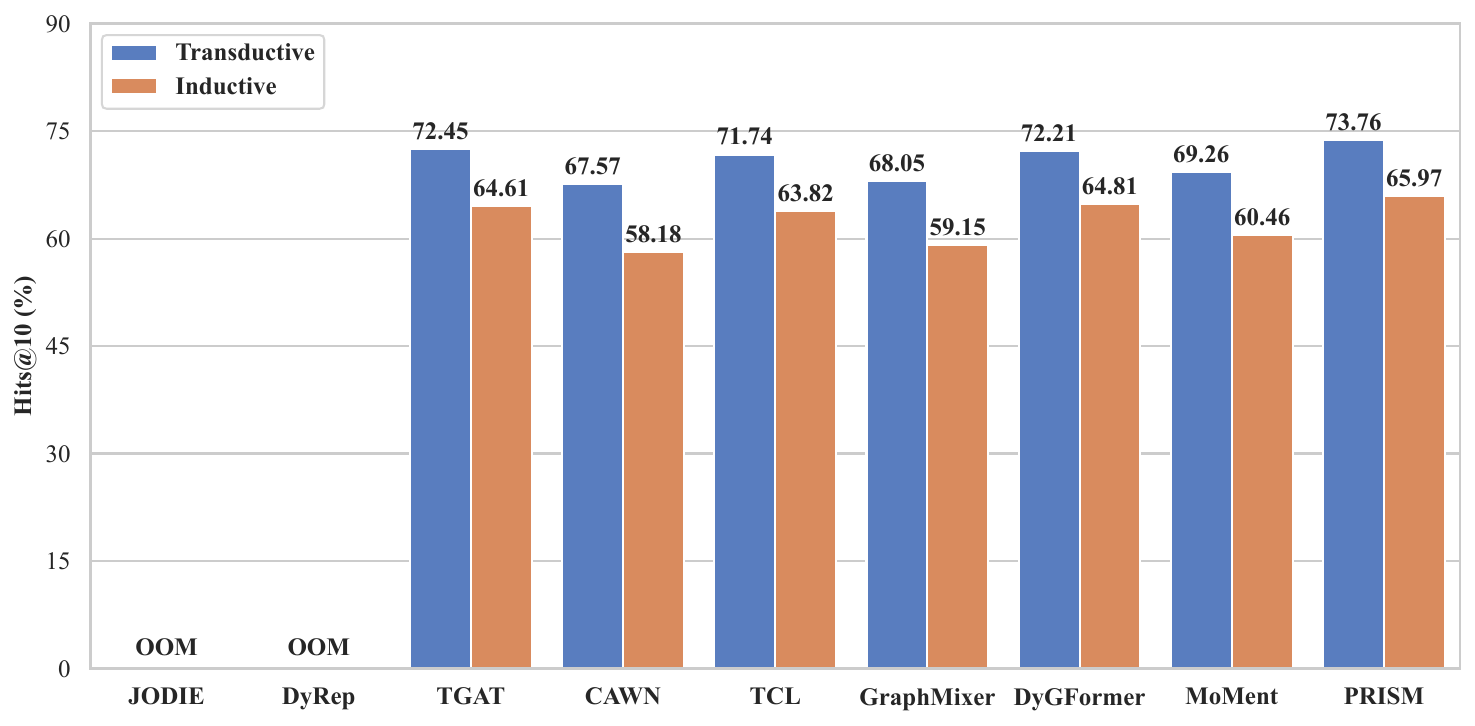}
        \caption{Amazon movies}
    \end{subfigure}
    \hfill
    \begin{subfigure}[b]{0.48\linewidth}
        \centering
        \includegraphics[width=\linewidth]{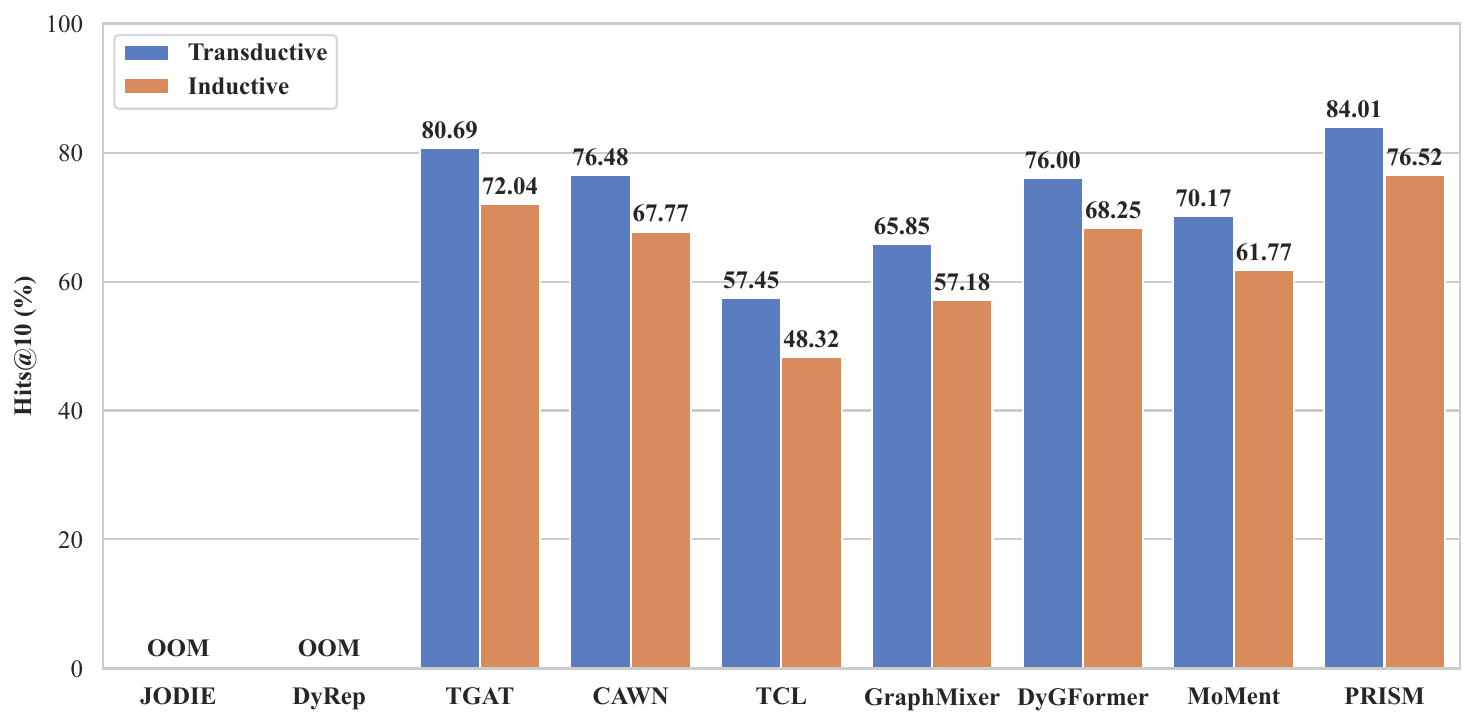}
        \caption{Yelp}
    \end{subfigure}
    \caption{Hits@10 (\%) results for destination node retrieval.}
    \label{fig:comp_result_hits10}
\end{figure}

Table~\ref{tab:comp_result_hits1} and Figure~\ref{fig:comp_result_hits10} report additional Hits@K results for destination node retrieval. Together with the Hits@3 results in the main text, these results show that PRISM maintains strong ranking performance across different retrieval thresholds. In particular, PRISM achieves the best average rank for both Hits@1 and Hits@10, further validating the effectiveness of semantic--behavioral posterior refinement for fine-grained destination ranking.

\begin{table}[htbp]
    \centering
    \caption{Hits@1 (\%) for destination node retrieval under transductive (\textit{tr.}) and inductive (\textit{in.}) settings. The best and the second best results are marked as \textbf{bold} and \underline{underlined}, respectively.}
    \label{tab:comp_result_hits1}
    \resizebox{\linewidth}{!}{
    \begin{tabular}{lcccccccccc}
    \toprule
     & \textbf{Datasets} & \textbf{JODIE} & \textbf{DyRep} & \textbf{TGAT} & \textbf{CAWN} & \textbf{TCL} & \textbf{GraphMixer} & \textbf{DyGFormer} & \textbf{MoMent} & \textbf{PRISM} \\
    \midrule
    \multirow{6}{*}{\textit{tr.}}
     & Enron
     & 50.45 $\pm$ 0.83
     & OOM
     & 53.66 $\pm$ 0.26
     & 65.90 $\pm$ 0.18
     & 43.44 $\pm$ 0.20
     & 44.05 $\pm$ 0.15
     & \textbf{74.73 $\pm$ 0.23}
     & 51.03 $\pm$ 0.35
     & \underline{73.23 $\pm$ 0.81} \\
     & ICEWS1819
     & 66.03 $\pm$ 0.10
     & 61.33 $\pm$ 0.26
     & 78.09 $\pm$ 0.31
     & 78.12 $\pm$ 0.51
     & \underline{81.88 $\pm$ 1.21}
     & 80.03 $\pm$ 0.78
     & 80.36 $\pm$ 0.02
     & 75.94 $\pm$ 0.16
     & \textbf{82.59 $\pm$ 0.21} \\
     & Googlemap CT
     & OOM
     & OOM
     & \textbf{22.90 $\pm$ 0.42}
     & 13.22 $\pm$ 0.19
     & 14.68 $\pm$ 0.04
     & 12.02 $\pm$ 0.05
     & 13.14 $\pm$ 0.02
     & 12.16 $\pm$ 0.02
     & \underline{21.38 $\pm$ 0.05} \\
     & GDELT
     & 28.90 $\pm$ 0.31
     & 31.12 $\pm$ 0.29
     & 41.35 $\pm$ 0.13
     & 42.53 $\pm$ 0.10
     & 43.53 $\pm$ 0.24
     & 38.75 $\pm$ 0.33
     & \underline{47.64 $\pm$ 0.08}
     & 43.10 $\pm$ 0.17
     & \textbf{50.49 $\pm$ 0.11} \\
     & Amazon movies
     & OOM
     & OOM
     & 28.51 $\pm$ 0.09
     & 22.37 $\pm$ 0.26
     & 28.41 $\pm$ 0.13
     & 26.13 $\pm$ 0.08
     & \textbf{30.08 $\pm$ 0.07}
     & 26.77 $\pm$ 0.97
     & \underline{29.97 $\pm$ 0.54} \\
     & Yelp
     & OOM
     & OOM
     & \underline{37.37 $\pm$ 0.51}
     & 32.18 $\pm$ 1.04
     & 15.54 $\pm$ 0.28
     & 22.04 $\pm$ 0.37
     & 35.12 $\pm$ 0.27
     & 27.16 $\pm$ 0.15
     & \textbf{44.40 $\pm$ 0.43} \\
     \midrule
     & \textbf{Avg. Rank}
     & 8.17 & 8.17 & 3.67 & 4.67 & 4.50 & 6.00 & \underline{2.50} & 5.83 & \textbf{1.50} \\
     \midrule
    \multirow{6}{*}{\textit{in.}}
     & Enron
     & 30.03 $\pm$ 0.42
     & OOM
     & 28.32 $\pm$ 0.33
     & 42.23 $\pm$ 0.29
     & 19.67 $\pm$ 0.14
     & 20.46 $\pm$ 0.09
     & \textbf{55.53 $\pm$ 0.27}
     & 24.56 $\pm$ 0.25
     & \underline{49.70 $\pm$ 0.13} \\
     & ICEWS1819
     & 51.34 $\pm$ 0.10
     & 47.99 $\pm$ 0.26
     & 57.52 $\pm$ 0.31
     & \underline{63.42 $\pm$ 0.51}
     & 60.59 $\pm$ 1.22
     & 61.14 $\pm$ 0.78
     & 63.40 $\pm$ 0.28
     & 54.35 $\pm$ 0.35
     & \textbf{66.79 $\pm$ 0.49} \\
     & Googlemap CT
     & OOM
     & OOM
     & \underline{5.50 $\pm$ 0.07}
     & 3.21 $\pm$ 0.04
     & 2.36 $\pm$ 0.05
     & 1.91 $\pm$ 0.01
     & 1.78 $\pm$ 0.06
     & 9.92 $\pm$ 0.04
     & \textbf{17.46 $\pm$ 0.18} \\
     & GDELT
     & 30.80 $\pm$ 0.74
     & 27.40 $\pm$ 0.51
     & 26.30 $\pm$ 0.42
     & 29.16 $\pm$ 0.11
     & \underline{31.99 $\pm$ 0.31}
     & 26.10 $\pm$ 0.46
     & \textbf{36.81 $\pm$ 0.38}
     & 17.66 $\pm$ 0.17
     & 28.86 $\pm$ 0.25 \\
     & Amazon movies
     & OOM
     & OOM
     & 24.03 $\pm$ 0.38
     & 18.35 $\pm$ 0.42
     & 24.63 $\pm$ 0.08
     & 22.24 $\pm$ 0.09
     & \textbf{26.13 $\pm$ 0.03}
     & 22.24 $\pm$ 0.98
     & \underline{25.23 $\pm$ 0.48} \\
     & Yelp
     & OOM
     & OOM
     & \underline{30.86 $\pm$ 2.38}
     & 26.42 $\pm$ 0.71
     & 13.83 $\pm$ 1.05
     & 18.41 $\pm$ 0.14
     & 28.98 $\pm$ 0.36
     & 22.50 $\pm$ 0.08
     & \textbf{37.03 $\pm$ 0.36} \\
     \midrule
     & \textbf{Avg. Rank}
     & 6.83 & 7.83 & 4.33 & 3.83 & 4.83 & 5.83 & \underline{2.50} & 7.00 & \textbf{2.00} \\
     \bottomrule
    \end{tabular}}
\end{table}

\end{document}